\documentclass[10pt,twocolumn,letterpaper]{article}

\usepackage{iccv}              % To produce the CAMERA-READY version
\usepackage{xcolor}
\usepackage{graphicx} % Required for inserting images
\usepackage{tcolorbox}
\usepackage{lipsum}
\usepackage{tabularray}
\usepackage[linesnumbered,ruled,vlined]{algorithm2e}
\usepackage{booktabs}
\usepackage{multirow}
\usepackage{leftidx}
\usepackage{bbm}
\usepackage{floatrow}

\usepackage{algorithmic}
\usepackage{textcomp}
\usepackage{siunitx}
\usepackage{tablefootnote}
\usepackage{footnote}
\usepackage{caption}
\usepackage{makecell}
\usepackage{pifont}% http://ctan.org/pkg/pifont

\usepackage{xspace}
\usepackage{amsmath}
\usepackage{amssymb}
\usepackage{graphicx}
\usepackage{floatrow}
\usepackage{wrapfig}

\newcommand{\myparagraph}[1]{\vspace{2pt}\noindent{\bf #1}}

\newcolumntype{P}[1]{>{\centering\arraybackslash}p{#1}}
\newcolumntype{M}[1]{>{\centering\arraybackslash}m{#1}}
\newcolumntype{L}[1]{>{\raggedright\arraybackslash}m{#1}}

\newcommand{\relativeimpP}[1]{\small{\textcolor{Green}{+#1}}}
\newcommand{\relativeimpN}[1]{\small{\textcolor{BrickRed}{-#1}}}
\newcommand{\relativeimpZ}[1]{\small{0.0}}

\newcommand{\relativeimpnP}[1]{\tiny{\textcolor{Green}{(+#1\%)}}}

\makeatletter
\def\blfootnote{\gdef\@thefnmark{}\@footnotetext}
\makeatother
\definecolor{iccvblue}{rgb}{0.21,0.49,0.74}
\usepackage[pagebackref,breaklinks,colorlinks,allcolors=iccvblue]{hyperref}

\usepackage{fontawesome5}
\usepackage{hyperref}

\makeatletter
\newcommand{\github}[1]{%
   \href{#1}{\faGithubSquare}%
}
\makeatother

\def\paperID{504} % *** Enter the Paper ID here
\def\confName{ICCV}
\def\confYear{2025}

\newcommand{\method}{Instructify\xspace}

\title{Instructify: Demystifying Metadata to Visual Instruction Tuning Data Conversion}

\author{
Jacob Hansen$^{\dagger 1}$ \and 
Wei Lin$^{\ddagger 2}$ \and
Junmo Kang$^3$ \and
Muhammad Jehanzeb Mirza$^4$ \and
Hongyin Luo$^4$ \and
Rogerio Feris$^5$ \and
Alan Ritter$^3$ \and 
James Glass$^4$ \and
Leonid Karlinsky$^{1}$ \\
$^1$Xero \ \ $^2$JKU Linz \ \ $^3$ Georgia Tech \ \ $^4$MIT CSAIL \ \ $^5$MIT-IBM Watson AI Lab \\
\github{} \url{https://github.com/jacob-hansen/Instructify} \\
}

\begin{document}

\twocolumn[{
\renewcommand\twocolumn[1][]{#1}
\maketitle
\centering
\vspace{-0.2cm}
\includegraphics[width=1.0\textwidth]{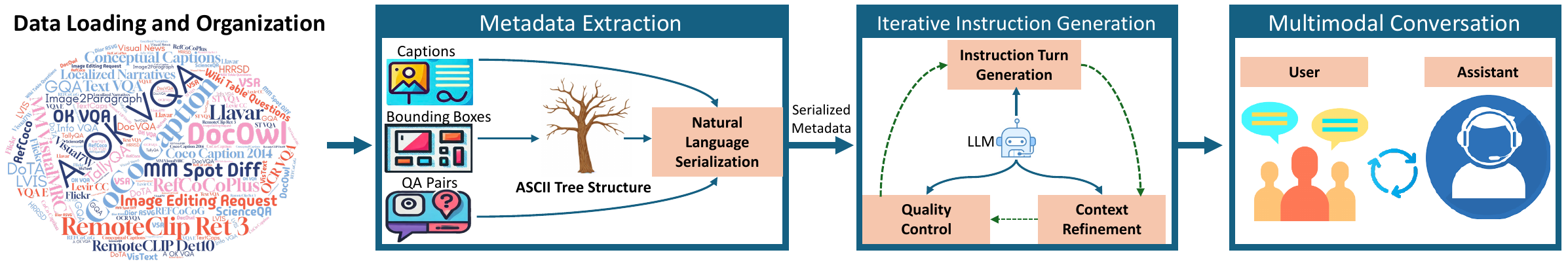}
\captionof{figure}{\textbf{Overview of \method.} 
% \wei{TODO: I feel that the caption does not align with the teaser figure ..  I would change "Natural Language Serialization" into "Serialized Metadata", "Conversation synthesis" into "Iterative Instruction-generation". LLM should be marked at the robot. "Context refinement" and "Quality control"   }
We present a unified framework that automatically transforms diverse metadata from publicly available datasets into multimodal instruction-tuning conversations. Our approach merges metadata from multiple sources, categorizing it into captions, bounding boxes, and question-answer (QA) pairs. While LLMs effectively convert captions and QA into natural language, we find that grounded annotations - such as bounding boxes - benefit significantly from a hierarchical representation using ASCII tree, capturing the geometric and semantic structure of the image. The serialized metadata then undergoes an iterative instruction-generation process, where context is refined and quality is rigorously controlled, with an LLM in the loop.
% We present a unified framework that converts variable metadata from an arbitrary combination of publicly available datasets into multi-modal instruction-tuning conversations in an automated manner. Most available metadata, automatically merged by us from multiple sources, can be categorized into captions, bounding boxes, and question-answer pairs (QA). While the conversion of captions and QA to natural language can be performed by Large Language Models (LLMs), we find that grounded annotations, such as the bounding boxes, significantly benefit from being converted into a hierarchical structure, an ASCII tree, representing the geometric and semantic organization of the image. Later, our natural language serialized metadata is fed to an iterative process of generating instruction turns, which are automatically refined through rigorous context fact reduction and quality control, again with an LLM in the loop.
% \wei{TODO R3 comments that the information conveyed in Figure 1 is inadequate and does not correspond well with the text in introduction that references it, necessitating further enhancement. The "Data loading and organization", "Information management" and "prompt management" mentioned are not clearly shown in Figure 1}
}
% \vspace{-0.1cm}
\label{fig:teaser}
\vspace{.1cm}
}]
\maketitle
{
\blfootnote{
$\dagger$ Work done during internship at MIT-IBM Watson AI Lab}
}
{
\blfootnote{
$\ddagger$ Correspondence: \tt\small{wlin2021at@gmail.com} }
}

\begin{abstract}
% We address the challenge of repeatability and scalability in synthetic dataset generation for visual instruction tuning. Unlike previous work aggregating diverse visual instruction datasets, often compromising reliability and safety, we propose a unified approach to optimize and standardize prompt strategies, sourced from many published works. Our platform enables rapid iteration on instruction fine-tuning, featuring an efficient framework for dataset control, prompt organization, and conversation sampling. Through comprehensive ablation studies, we explore the impact of multiple factors, including conversation format, base model selection, resampling strategies, and dataset scaling. Our experiments demonstrate the feasibility of surpassing GPT-4-generated instructions by leveraging weaker models—such as Gemma 2 27B, LLaMa 3.1 70B, and Qwen 2.5 72B—achieving a 3-8\% improvement across a range of benchmarks, including <INSERT BEST 1>, <INSERT BEST 2>, and <INSERT BEST 3>. Furthermore, we provide evidence of our system’s adaptability and scalability through its application to the llava-1.5 and Cambrian datasets, where we either replace GPT-4-generated data or maximize the use of resampled conversations to enhance model performance. Ultimately, our platform advances open-source visual instruction tuning by reducing dependency on closed-source models, thus supporting the development of specialized instruction tuning datasets for niche domains.

Visual Instruction Tuning (VisIT) data, commonly available as human-assistant conversations with images interleaved in the human turns, are currently the most widespread vehicle for aligning strong LLMs to understand visual inputs, converting them to strong LMMs. 
While many VisIT datasets are available, most are constructed using ad-hoc techniques developed independently by different groups. They are often poorly documented, lack reproducible code, and rely on paid, closed-source model APIs such as GPT-4, Gemini, or Claude to convert image metadata (labels) into VisIT instructions. 
% This incurs significant cost and difficulty to scale, improve quality, or produce VisIT data for new datasets. 
This leads to high costs and makes it challenging to scale, enhance quality, or generate VisIT data for new datasets.
In this work, we address these challenges and propose an open and unified recipe and approach,~\textbf{\method}, for converting available metadata to VisIT instructions using open LLMs. Our multi-stage \method features an efficient framework for metadata grouping, quality control, data and prompt organization, and conversation sampling. We show that our approach can reproduce or enhance the data quality of available VisIT datasets when applied to the same image data and metadata sources, improving GPT-4 generated VisIT instructions by ~3\% on average and up to 12\% on individual benchmarks using open models, such as Gemma 2 27B and LLaMa 3.1 70B. 
Additionally, our approach enables effective performance scaling - both in quantity and quality - by enhancing the resulting LMM performance across a wide range of benchmarks. We also analyze the impact of various factors, including conversation format, base model selection, and resampling strategies. 
Our code, which supports the reproduction of equal or higher-quality VisIT datasets and facilities future metadata-to-VisIT data conversion for niche domains, is released at \url{https://github.com/jacob-hansen/Instructify}. 
% We further show that our approach enables effective performance scaling (in terms of resulting LMM performance on a large variety of benchmarks) of the produced VisIT data both in terms of quantity and quality. 
% Additionally, we explore the impact of multiple factors, including conversation format, base model selection, and resampling strategies. 
% Our code, enabling reproducing equal or better quality existing VisIT datasets, as well as future expansion of metadata to VisIT data conversion for niche domains, will be released upon acceptance.

\end{abstract}
\section{Introduction}
\label{sec:intro}

% \wei{TODO  Reviewer2 mentioned that the novelty of the method is not framed properly. The key contribution is not just the use of open-source LLMs for data generation, but rather, as noted by the reviewers, the identification of a robust approach for creating data comparable to that generated by proprietary models. Including more details, if not in the main paper, then at least in the supplementary, would be essential for acceptance.}

Large Multimodal Models (LMMs) have enabled significant advancements, allowing billion-parameter-scale language models to effectively perform multimodal tasks such as comprehensive image captioning~\cite{blip2,wang2023caption}, visual question answering~\cite{instructblip,llava,llava-ov} and complex visual reasoning~\cite{llava, llava-ov,minigpt,mplug-owl,chen2023internvl}. 

The advancement of LMMs, however, has increasingly relied on a distillation paradigm, wherein high-performing proprietary systems, such as GPT-4~\cite{openai2023gpt}, are leveraged to generate training data through carefully designed prompts. 
While these systems enable powerful visual understanding and instruction-following capabilities, the field's reliance on closed-source models introduces substantial limitations. The top-performing open-weight LMMs have increasingly used data that is either closed-source~\cite{zhang2023internlm,flamingo} or collected by commercial models~\cite{llava,zhao2023svit,sharegpt4v,sharegpt4video,lrv_instruct}. This creates a landscape where transparency and reproducibility are restricted. 
% In this case, strongest open-weight LMMs rely on either generation on commercial models~\cite{llava,zhao2023svit,sharegpt4v,sharegpt4video,lrv_instruct} or inaccessible data sources. This reliance not only raises issues related to transparency, cost, and access but also poses substantial obstacles in repeatability, generalizability and scalability. 
This reliance on proprietary data pipelines leads to several core challenges: (1) \textit{Repeatability} suffers due to the lack of open, accessible code and documentation for data collection, making it difficult to replicate results accurately. (2) \textit{Generalizability} is limited as existing methods often rely on dataset-specific configurations, reducing their applicability across diverse domains. (3) \textit{Scalability} is hindered by the high costs and restrictive access associated with proprietary APIs. 
Existing approaches to Visual Instruction Tuning (VisIT) data generation are largely ad-hoc, developed by individual groups with minimal standardization. These methods often rely on closed-source model APIs like GPT-4~\cite{openai2023gpt}, Gemini~\cite{team2023gemini}, or Claude~\cite{claude} to convert image metadata into VisIT instructions~\cite{sharegpt4v,sharegpt4video,zhao2023svit,lvis_instruct,allava}, leading to high costs and scalability issues.

% Specifically, the \textit{repeatability} is impeded by the opaqueness of closed systems, making it difficult for researchers to replicate results. The \textit{generalizability} is limited as existing methods often rely on dataset-specific configurations, reducing their applicability across diverse domains. The \textit{scalability} is essential for the advancement of LMMs, yet efficient large-scale data generation remains challenging under current proprietary-focused methodologies. 

In this paper, we present a complete, open-source platform which we name \method, designed to overcome these limitations, enabling high-quality, scalable, and reproducible data generation for visual instruction tuning. 
An overview is provided in Figure~\ref{fig:teaser}. 
% \wei{introduce our method  - is the description correct? refer to figure?}
Specifically, our Instructify consists of four primary steps. First, \textit{Data Loading and Organization} manages a wide range of datasets and groups metadata by image source. \textit{Metadata Conversion} convert the metadata of different types into textual descriptions. \textit{Information Management and Quality Control} enhances data quality and diversity by filtering redundant information and employing LLM-based quality checks. Eventually \textit{Prompt Management} dynamically selects prompt templates, adjusting for different instruction styles and task requirements based on the metadata context. 

Our contributions are threefold: (1) we introduce a unified, open-source framework that consolidates metadata across diverse datasets and standardizes prompt engineering strategies, thereby enhancing \textit{replication} and \textit{accessibility}. (2) Our system is architected for \textit{scalability}, achieving vertical scaling through a high-throughput backend and horizontal scaling using a file-based control system for distributed processing. (3) We reduce reliance on closed-source models by demonstrating that high-quality synthetic data can be generated with open models, such as Gemma 2 27B~\cite{gemma2} and LLaMA 3.1 70B~\cite{llama3_1}, achieving comparable or superior quality to data generated by proprietary systems like GPT-4. 
% \wei{TODO R3 mentions that this paragraph can be integrated into the last paragraph}
Our extensive experiments show that our \method not only matches but often enhances the data quality found in existing VisIT datasets, improving GPT-4-generated instructions by ~3\% on average (over 12 benchmarks) and up to 12\% on individual benchmarks. Furthermore, we explore the impact of various factors, including conversation format, base model selection, and resampling strategies, on data quality and performance across benchmarks. By providing a reproducible, adaptable and scalable solution, our work lays a foundation for advancements in visual instruction tuning. Our codebase, allowing for the reproducibility of existing datasets as well as expansion to niche domains, will be open-sourced for open, transparent research in LMM development.

\section{Related Work}

\myparagraph{Large Multimodal Models (LMMs).} 
Both academia and industry have made significant strides in developing Large Multimodal Models (LMMs) that aim to match the multimodal reasoning capabilities of top-performing proprietary models such as GPT-4o~\cite{openai2023gpt4}, Gemini~\cite{team2023gemini}, and Claude~\cite{claude}. 
The LLaVA series~\cite{llava, llava+, llava-next} introduces a framework for multimodal instruction tuning, leveraging vision-language alignment through instruction-following data. These models incrementally expand the number of visual tokens during training to enhance vision-related reasoning and image comprehension. Math-LLaVA~\cite{math-llava} extends this approach to mathematical reasoning, integrating a specialized dataset, MathV360K, and fine-tuning a LLaVA-1.5-based model~\cite{llava+} to improve performance on mathematical benchmarks.
VILA2~\cite{vila2} focuses on iterative self-refinement, where the model continuously re-annotates its own pretraining data, retraining from scratch with these refined annotations to enhance performance. This method contrasts with traditional approaches that rely on external proprietary models or large-scale web crawled data. Molmo~\cite{molmo} advances multimodal grounding by incorporating high-quality human-annotated descriptive captions and a dataset of in-the-wild question-answer pairs, combined with 2D pointing data to connect language with spatial understanding in images.
Cambrian-1~\cite{cambrian1} introduces a vision-centric approach that bridges visual instruction tuning with visual representation learning. By evaluating over 20 vision encoders within a unified LMM training framework, it highlights the importance of data balance and source distribution for multimodal model performance.
In contrast to these approaches, our work focuses on optimizing instruction tuning by repurposing existing vision-language datasets. We introduce a unified framework that automates data aggregation, restructuring, and filtering to create high-quality instruction-tuning conversations, improving multimodal learning efficiency without relying on proprietary models or large-scale manual curation.

\myparagraph{Data Collection for Multimodal Instruction Tuning.} 
A common approach in multimodal instruction tuning is to repurpose existing task-specific datasets. Numerous Large Multimodal Models (LMMs) employ data collection pipelines that adapt high-quality samples from these datasets into instruction-following data. Models such as InstructBLIP~\cite{instructblip}, X-LLM~\cite{X-llm}, LLaMA-Adapter~\cite{llama-adapter}, Vision-LLM~\cite{wang2024visionllm}, LaVIN~\cite{luo2024cheap}, MultiInstruct~\cite{multiinstruct}, LLaMA-Adapter v2~\cite{llama-adapter-v2}, and ChatBridge~\cite{chatbridge} transform these samples into instruction-response pairs, with instructions either manually designed or generated by models like GPT. Many works prepare candidate instruction sets and randomly sample from them during training, either drafting the instructions manually~\cite{minigpt, llava-med, instructblip, X-llm, multiinstruct, m3it} or using a manually curated seed set to prompt GPT for additional instructions~\cite{videochat, wang2024visionllm, multimodal-gpt}.

Early efforts in multimodal instruction tuning focused on generating conversations by prompting proprietary LLMs. It can be viewed as a form of distillation from high-performing commercial models. For example, LLaVA~\cite{llava} constructs its instruction dataset, LLaVA-Instruct-150K, by using manually annotated captions and bounding boxes as symbolic representations of images and prompting text-only GPT-4 to generate multi-round instruction-response pairs, including simple QA, detailed descriptions, and complex reasoning tasks. SVIT~\cite{zhao2023svit} scales this approach further, using GPT-4 to generate 4.2M instruction-following samples, incorporating conversation QA, complex reasoning, referring QA, and detailed image descriptions, leveraging image annotations from Visual Genome~\cite{visual_genome}.

The introduction of GPT-4V has led to a new wave of data collection methodologies centered around vision-language generation. ShareGPT4V~\cite{sharegpt4v} compiles 100K high-quality descriptive captions for images from LAION~\cite{openclip}, CC~\cite{conceptual_captions}, SBU~\cite{sbu}, and COCO~\cite{microsoftcoco}, along with 200K instruction-following samples. Expanding on this, ShareGPT4Video~\cite{sharegpt4video} collects 40K dense video captions generated by GPT-4V. LVIS-Instruct4V~\cite{lrv_instruct} contributes 220K visually aligned, context-aware instructions using images from LVIS~\cite{lvis}. ALLaVA~\cite{allava} further extends this with 1.3M fine-grained image captions and visual QA pairs generated by GPT-4V.
To mitigate reliance on proprietary models, we adopt an open-source approach, replicating the data collection methodologies traditionally dependent on closed-source systems. Our work demonstrates that high-quality instruction-tuning data can be generated using open models while maintaining strong performance, offering a scalable and reproducible alternative for multimodal learning.

% \wei{}

% \wei{Data Collection Without Reliance on Proprietary Models.} Molmo - manual annotation of dense caption via narration

% \wei{Self-refinement,  self-augmentation}

% \wei{Mixture of Text-Only and Multimodal Data.} It is common practice to combine language-only conversation data and multimodal instruction-tuning data for enhancing the instruction-following capabilities of the LMMs~\cite{mplug-owl,multimodal-gpt,luo2024cheap,llama-adapter-v2}. The training batches are constructed either by random sampling~\cite{luo2024cheap} or sequential sampling~\cite{multiinstruct}. 

\myparagraph{Quality Enhancement of VisIT Data.} 
Instruction-tuning experiments in large multimodal model research consistently demonstrate that data quality often outweighs data quantity in driving model performance~\cite{zeng2024matters, wei2023instructiongpt}. Models such as Flamingo~\cite{flamingo} and LLaVA~\cite{llava} leverage carefully curated datasets that integrate diverse visual and textual components, enhancing the alignment between complex instructions and accurate visual understanding. Studies have further shown that incorporating diverse scene compositions, object relationships, and multi-step reasoning tasks leads to more robust learning~\cite{zhao2023svit, llava-onevision, math-llava}. In our work, we enhance VisIT data quality by transforming complementary annotations—such as QA pairs, abstract information, and bounding boxes associated with the same visual data—into structured textual descriptions, ensuring greater diversity in the generated data. Additionally, we implement rigorous information management and quality control mechanisms to maintain high data integrity throughout the process. Extensive experiments confirm that models fine-tuned on our curated instruction-tuning data outperform those trained on the originally proposed datasets, demonstrating the effectiveness of our approach in improving multimodal learning.

% Instruction-tuning experiments in numerous large multimodal model studies consistently highlight that data quality often has a greater impact than data quantity~\cite{zeng2024matters,wei2023instructiongpt}. 
% % Factors such as prompt diversity, task coverage and data curation are proved essential in data quality. 
% Works such as Flamingo~\cite{flamingo} and LLaVA~\cite{llava}  leverage curated datasets that incorporate varied visual and textual elements, bridging the gap between complex instructions and accurate visual understanding. Research has shown that incorporating diverse scene compositions, object relationships and multi-step instructions contributes more robust learning~\cite{zhao2023svit,llava-onevision, math-llava}. 
% In our work, we transform existing complementary annotations—such as QA pairs, abstract information, and bounding boxes associated with the same visual data—into textual descriptions to ensure high diversity in the generated data. Additionally, we implement information management and quality control procedures to maintain data quality throughout the process.
% Through extensive experiments we find that our curated instruction tuning data can help fine-tune models, which show improved performance, in comparison to fine-tuning on the originally proposed data.

% Math-LLaVA demonstrates the significance of dataset diversity and synthesis .

% \wei{data quality and data curation. }

% \wei{ }

\begin{figure*}[ht]
% \vspace{-0.3cm}
\centering
\vspace{-0.1cm}
\includegraphics[width=1\linewidth]{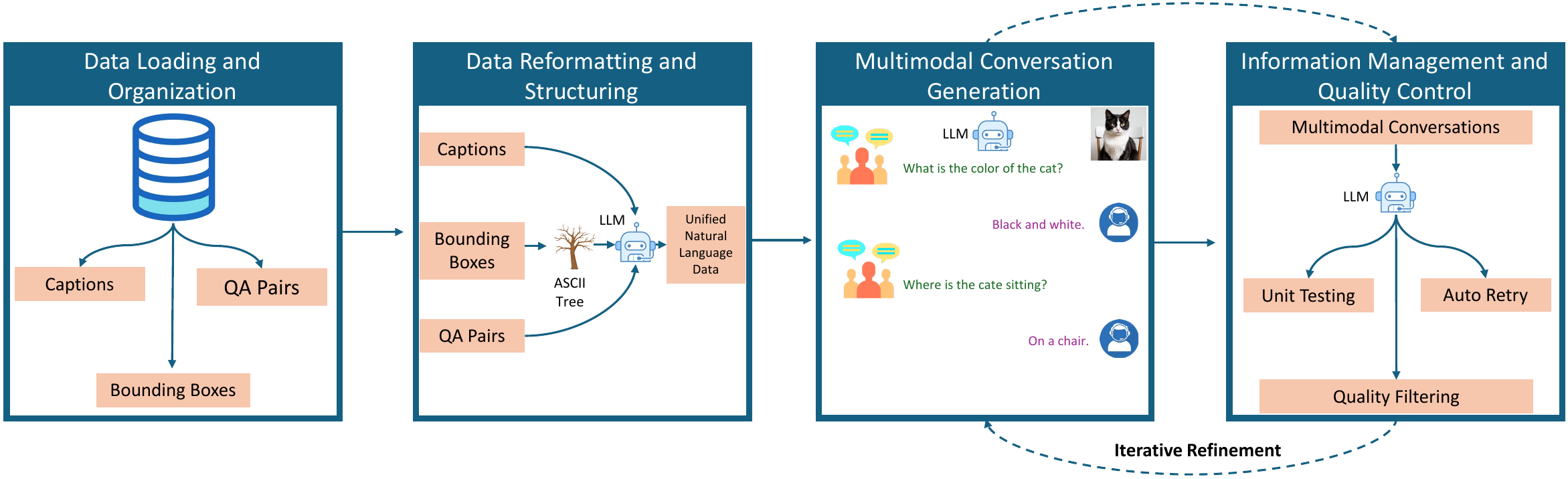}
\caption{
\label{fig:pipeline}
\textbf{Pipeline of our method.} We start with loading and organizing multiple open-source datasets containing captions, question answers, and bounding boxes as metadata. Next, we reformat the metadata in a unified natural language interface which is then converted into multi-modal conversations. 
% The conversations can be erroneous as they are generated automatically through an LLM. 
These LLM-generated conversations undergo an iterative refinement procedure involving multiple automated tests for quality improvement.  
% \wei{(Reviewer comment: the flow of information is not always clear in the diagrams. E.g. where does the "Unified natural language data" in the diagram go?
% )\\
% R3 mentions that we should modify Figure 2 by incorporting examples of the corresponding technical details. This would enhance the figure's informational value and aid readers in comprehending the methodology. 
% }
}
\vspace{-0.3cm}
\end{figure*}

\section{\method}
\label{sec:method}

% \wei{TODO  R3:  Ambiguities within the methodology section may hinder the community's comprehension of the method's novelty. }
% Our proposed pipeline \method addresses the challenges of high-quality visual instruction tuning data generation. We offer a flexible pipeline that transforms diverse image metadata aggregated automatically from existing datasets into high-quality visual instruction (human-assistant) conversations. We introduce several innovations across data processing, metadata organization into LLM prompts, automatic data quality control, and conversation generation that enable efficient scaling and reproducibility.

% \subsection{System Overview}
% \wei{TODO  R3 thinks that Section 3.1 is too long }

% \wei{TODO: add a pipeline figure of system architecture, with an example for each component in the architecture, e.g.  the 3 types of data in the loading step,  how bounding boxes are converted into captions by the conversion system, how the prompt templates are randomly sampled, etc.  }

% Our approach pipeline for transforming image metadata into instruction-tuning conversations consists of four primary components elaborated in the following. Our approach is illustrated in Fig. \ref{fig:pipeline} and our code is provided in the Supplementary.

We introduce a flexible pipeline that converts diverse image metadata into high-quality visual instruction conversations through four key components: (1) \textbf{Data Loading and Organization} manages over 40 public datasets, grouping metadata by image source and reformatting QA data into factual statements for better generalization; (2) \textbf{Metadata Conversion} transforms various annotations (e.g., bounding boxes, captions, QA) into structured text, introducing an ASCII tree representation to capture object attributes and relationships; (3) \textbf{Information Management and Quality Control} ensures instruction diversity and accuracy through automatic fact reduction, LLM-based filtering, and quality checks; (4) \textbf{Prompt Management} dynamically selects prompt templates, mixing instruction styles and task intents based on metadata compatibility. An overview of our approach is shown in Fig. \ref{fig:pipeline}, with code provided in the Supplementary.

%%%%% data loading and organization
\subsection{Data Loading and Organization}\label{sec:data_loading} 
The Data Loading and Organization component of our approach provides a comprehensive framework for managing and unifying diverse visual-language datasets with a variety of supported metadata formats. Our implementation currently supports over 40 datasets, including caption datasets (e.g., COCO Captions~\cite{microsoftcoco}, Localized Narratives~\cite{localized_narratives}, Flickr30K~\cite{flickr30k}), object reference collections (e.g., Visual Genome~\cite{visual_genome}, LVIS-Instruct4V~\cite{lvis_instruct}, RefCOCO~\cite{refcoco}), and visual question-answering datasets (e.g., GQA~\cite{gqa}, OK-VQA~\cite{okvqa}, DocVQA~\cite{docvqa}). We establish metadata grouping by image source to facilitate metadata merging, handling three fundamental types of input data: \textit{bounding boxes with associated attributes}, \textit{direct image captions}, and \textit{question-answer (QA) pairs}. Importantly, we reformat automatically extracted QA data into factual statements, which improves generalization for novel conversation generation and prevents repetitive QA pairs.

While our implementation builds upon HuggingFace Datasets for core functionality, it also extends support to datasets that cannot be hosted on public platforms due to licensing or permission constraints. Our data loading interface flexibly allows new datasets to be integrated with minimal effort. By implementing a small set of methods, users can incorporate additional datasets, enhancing the diversity of available data and enabling extensions to specialized imagery or additional metadata annotations.

Each dataset is encapsulated in a dedicated Python class that manages dataset-specific directory structures, annotation formats, and data splits, exposing a standardized set of methods through our generic interface. Our approach supports cross-dataset metadata grouping through the automatic identification and linking of shared images across collections. When an image appears in multiple datasets, our system aggregates all available annotations while preserving their original source provenance. This capability enables the creation of rich training instances by combining complementary information types, such as merging detailed object annotations from Visual Genome with natural language descriptions from COCO Captions.

%%%%%%%%%%%%% metadata formatting and conversion 
\subsection{Metadata Formatting and Conversion}\label{sec:metadata_formatting}
The core insight of our metadata formatting pipeline is that diverse types of image metadata labels can be effectively unified through conversion to a structured textual form. While existing approaches maintain separate processing paths for different data types (bounding boxes, QA pairs, captions), we show that converting all information into a unified structured format enables more flexible downstream processing and better generalization. This unified representation effectively conveys the structured organization of image context to the instruction-synthesizing LLM, leading to more consistent prompt performance, simplified quality control, and higher-quality, more diverse visual instructions. Our system processes three fundamental types of metadata from our diverse set of supported datasets: question-answer (QA) pairs, bounding boxes with attributes and direct image captions. 

\myparagraph{QA Pairs and Abstract Information.}
We transform QA pairs into declarative statements using an LLM-based prompt that preserves semantic content while removing the question-answer structure. For example, the QA pair \textit{Q: What color is the car? A: Red} is reformatted as \textit{There is a red car in the image.} This approach ensures that the core information is retained while making it more suitable for integration into natural conversations. Additionally, this conversion prevents biasing the model toward simply repeating QA pairs, improving its ability to generate novel and contextually appropriate responses.

 \begin{figure*}[ht]
% \vspace{-0.3cm}
\centering
\vspace{-0.2cm}
\includegraphics[width=1\linewidth]{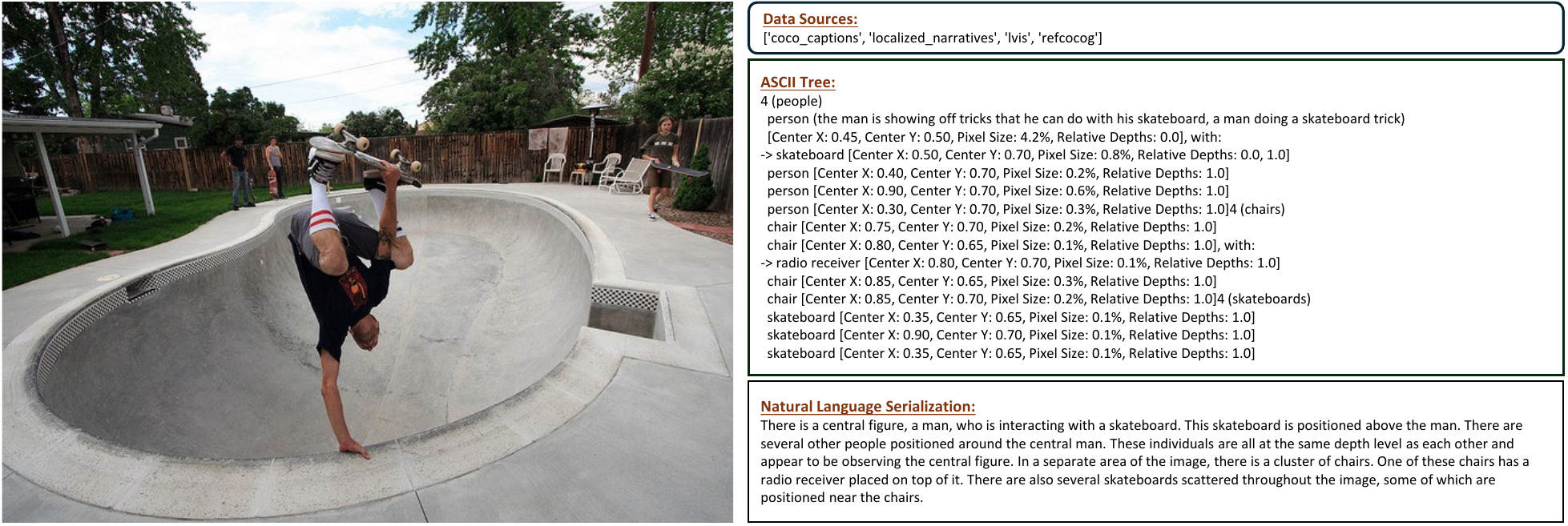}
\caption{
\label{fig:ascii_tree}
\textbf{ASCII tree example.} Our approach automatically collects all available metadata from the available sources (top right) for the image (left). As detailed in the text, we propose a way to re-organize this information into a hierarchical textual data structure - the ASCII tree (middle right) that is effectively converted by an LLM into a detailed high-quality context comprised of multiple factual statements organized in sentences, densely describing the image objects, their relations, and attributes. The resulting context is later fed into our iterative instruction turn generation process detailed in the text.
% \wei{(TODO  The reviewers asks for the explanation of the term "Tree sterilization")}
}
\vspace{-0.3cm}
\end{figure*}

\begin{algorithm}[t!]
% \small
\scriptsize
\DontPrintSemicolon
\caption{ASCII Tree Builder}
\label{alg:hierarchical}

% Brief variable definitions
\KwIn{
    $I$: image,
    $B$: bounding boxes,
    $t_s$: spatial threshold,
    $t_m$: mask threshold,
    $t_c$: containment threshold
    % \wei{R3 asks why the $t_s$ and $t_c$ are not used in the pseudo code}
}
\KwOut{$T$: scene hierarchy tree}

\Begin{
    $B.compute\_masks()$\tcp*{SAM2\cite{ravi2024sam2segmentimages}}
    $B.compute\_depth()$\tcp*{Depth Anything v2\cite{yang2024depthv2}}
    $R = B.merge()$\;
    
    \textcolor{teal}{Build Tree} \;
    initialize $tree$
    \While{$R \neq \emptyset$}{
        $r = R.pop\_max()$\;
        $tree.add(r)$\;
        $tree.add(r.contained(t_c))$
    }
    
    \textcolor{teal}{Format Tree}\;
    \SetKwFunction{FF}{F}
    \SetKwProg{Fn}{Function}{:}{}
    \Fn{\FF{$node, intent$}}{ 
        $node.group()$\tcp*{matching labels under new child}
        $node.sort()$\;
        $s = node.format(d, indent)$
        \For{$child$ \textbf{in} $node$}{  
            $s += F(child, intent + 1)$
        }
        \Return{$s$}\;
    }
    F(tree.root)
}
\vspace{-0.2cm}
\end{algorithm}

% \end{document}

\begin{algorithm}[hbt!]
% \small
\scriptsize
\DontPrintSemicolon
\caption{Visual Instruction Generation}
\label{alg:instruction_gen}
\KwIn{
    $\mathcal{S}$: full context sentences \\
    $r_t$: reduction threshold (0.85) \\
    $l_{min}$: minimum information length (100)\\
    $P$: prompt distribution 
    %\wei{R3 mentions that prompt distribution is unclear in the method section}
}
\KwOut{$R$: generated QA pairs}
\Begin{
    $\mathcal{S}_0 = \mathcal{S}$  \tcp*{Initial context}
    $R = []$  \tcp*{Results}
    $i = 0$  \tcp*{Iteration counter}
    
    \While{not StoppingCriteria()}{
        \textcolor{teal}{Generate Instructions}\;
        $p = SamplePrompt(P)$ 
        %\wei{TODO R3 mentions that the techinical details of SamplePrompt is not clear} \; 
        $qa = GenerateQA(\mathcal{S}_i, p)$\;
        
        \textcolor{teal}{Verify Against Full Context}\;
        \If{$VerifyQA(qa, \mathcal{S})$}{
            $R.append(qa)$\;
            
            \textcolor{teal}{Filter Used Information}\;
            $\mathcal{S}_{i+1} = Reduce(\mathcal{S}_i, qa)$\;
            $i = i + 1$\;
        }
    }
    \Return{$R$}
}

\tcp{Stopping criteria function}
\SetKwFunction{FSC}{StoppingCriteria}
\SetKwProg{Fn}{Function}{:}{}
\Fn{\FSC{}}{
    $len\_ratio = len(\mathcal{S}_i) / len(\mathcal{S})$\;
    \Return{$len\_ratio < (1 - r_t)$ or $len(\mathcal{S}_i) < l_{min}$}
}
\vspace{-0.2cm}
\end{algorithm}
\vspace{-0.2cm}

\myparagraph{Bounding Boxes with Associated Attributes.}
Converting bounding box annotations into natural language descriptions is a more complex challenge, which we address through our proposed hierarchical ASCII tree representation. This structured textual format captures objects, their attributes, and spatial and inter-object relationships, effectively preserving scene composition in an LLM-friendly manner. 
The process begins with the construction of an ASCII tree that represents the spatial and semantic relationships between objects. Tree nodes correspond to image objects, object groups, or object parts, and are enriched with attributes such as center coordinates and pixel dimensions. Containment relationships (e.g., object-part, object-group) are determined using bounding box and segmentation mask intersection criteria, with strict thresholds applied for merging similar objects and more relaxed thresholds for establishing containment relationships.
Additionally, depth information, extracted using Depth Anything V2~\cite{yang2024depthv2}, is incorporated to ensure that objects with different depth levels are not mistakenly merged. 

The tree structure is then refined by grouping and sorting objects based on depth, mask size, and location. A counting mechanism distinguishes between countable and uncountable objects, using descriptors like “many” or “several” for large groups instead of exact numerical counts. 
% \wei{TODO Here, R3 mentions that we have ambiguities in the method description. we simply skim over the handling of groups of identical objects and groups of different objects. While authors provide the code in the supplementary, such critical technical details should still be explicitly addressed in the main text. }
For groups of identical objects, the system generates concise descriptions using averaged measurements. In more complex cases, where objects have distinct characteristics or hierarchical relationships, the ASCII tree format preserves these details through structured indentation. This hierarchical organization ensures that the final output remains both human-readable and LLM-friendly, allowing for natural and rich spatial descriptions. An example of an image and its corresponding tree structure, automatically generated from multiple metadata sources, is illustrated in Fig.~\ref{fig:ascii_tree}, while the pseudo-code for metadata merging into a tree structure is provided in Algo.~\ref{alg:hierarchical}.

\myparagraph{Direct Image Captions.}
Beyond the structured tree representation, our system seamlessly incorporates existing image captions directly from metadata sources, ensuring that all available descriptive information is utilized for enhanced instruction generation.

%%%%%%%%%%%%%%% Information management, quality control and instruction generation 
\subsection{Information Management, Quality Control, and Instruction Generation}\label{sec:info_management}
Our generated visual instructions consist of human-assistant conversation turns, progressively constructed in stages starting from a comprehensive set of context sentences, denoted as $\mathcal{S}$. To ensure instruction diversity without compromising quality, we implement an automatic filtering mechanism that reduces redundancy and maximizes metadata utilization. At each stage $i$, the current set of remaining context sentences $\mathcal{S}_i$ is evaluated against the already-generated conversation turns $\mathcal{C}_i$ to identify and remove redundant or previously used information. This strategy prevents excessive repetition, ensuring that generated instructions explore different aspects of an image and contribute to a more varied dataset. To balance diversity and completeness, we terminate conversation generation once 85\% of the original metadata has been incorporated into instructions.
% \wei{(TODO  Here reviewer asks wether 85\% of removal threshold was calculated empirically.  )}
The filtered context $\mathcal{S}_i$ is then input into an LLM along with an instruction generation prompt, which is randomly sampled from our prompt management system (detailed in Sec.~\ref{sec:prompt_management}). This ensures that different styles and task intents are systematically applied to the instruction generation process. To maintain factual correctness and prevent hallucinations, the original full context $\mathcal{S}$ is retained and used for verification. Instructions generated from the filtered context $\mathcal{S}_i$ are cross-checked against $\mathcal{S}$ to detect inconsistencies or contradictions. If factual errors are identified, the system triggers an auto-retry mechanism that prompts the LLM to regenerate the conversation turn, refining it until it aligns with the original metadata.

The instruction generation process concludes when either (i) the remaining context contains insufficient information (fewer than 100 characters) or (ii) the system has removed 85\% of the original metadata. Additionally, automatic LLM-based quality filtering and unit testing are integrated to enforce instruction clarity, correctness, and coherence. The complete algorithm for context filtering and instruction generation is provided in Algo.~\ref{alg:instruction_gen}. 
% \wei{TODO R3 mentions that the details of unit-testing and auto-entry are not clear.  R2 mentions that more details on the algorithm for identifying and removing "used" information are needed. }
% \vspace{-0.2cm}
%%%%%%%%%%%%%%%%%%%%%%% Prompt management 
\subsection{Prompt Management}\label{sec:prompt_management}
% \wei{TODO  R3 mentions the technical details of the "dynamic sampling of prompts" are not clear}
The prompt management component implements a modular system for dynamically handling diverse prompt strategies. It is responsible for sampling and applying various instruction styles and downstream task intents while ensuring compatibility with the available converted metadata context. The system takes as input a list of captions and predefined prompt templates, selecting templates according to configurable probability distributions. This approach ensures a balanced mix of different instruction styles, promoting variation in the generated conversations.

To extract structured human-assistant dialogues, the system applies regular expression-based parsing to the generated output. If no valid conversation is obtained, an error-handling mechanism triggers up to three automatic retries before moving on. This ensures robust instruction generation while minimizing failures due to incomplete or incompatible responses. By systematically varying prompt structures and sampling strategies, the prompt management component enhances the diversity, adaptability, and contextual alignment of the generated visual instructions.

\section{Experiments}

% \begin{table}[ht]
% \begin{tabular}{ccccccccccccccc}
% \toprule
% Benchmarks & LLaVA Mix & LLaVA Mix with replacement \\
% \midrule
% GQA        &           &                            \\
% TextVQA    &           &                            \\
% ChartVQA   &           &                           \\
% \bottomrule
% \end{tabular}
% \caption{This is a dummy table.}
% \label{tab:table1}
% \end{table}

\begin{table*}[ht]
\scriptsize
\begin{tabular}{l|cc|cc|cc}
\toprule
    Model         & LLaVA   & Our LLaVA & PF-1M   & Our PF  & LLaVAR  & Our LLaVAR \\
   \# Samples & 160K    & 160K      & 1M      & 602K    & 16K     & 16K        \\
\midrule
\midrule
AI2D~\cite{ai2d}         & 62.82\% & 63.18\% (\relativeimpP{0.36\%})  & 61.59\% (\relativeimpN{1.23\%}) & 63.05\% (\relativeimpP{0.23\%})& 62.53\% (\relativeimpN{0.29\%})& 63.31\%  (\relativeimpP{0.49\%})  \\
ChartQA~\cite{masry-etal-2022-chartqa}      & 17.68\% & 19.44\% (\relativeimpP{1.76\%})  & 19.44\% (\relativeimpP{1.76\%}) & 20.28\% (\relativeimpP{2.60\%})& 19.44\% (\relativeimpP{1.76\%})& 18.76\%  (\relativeimpP{1.08\%})  \\
DocVQA\textsubscript{val}~\cite{docvqa}   & 45.41\% & 45.43\% (\relativeimpP{0.02\%})  & 46.78\% (\relativeimpP{1.36\%}) & 49.93\% (\relativeimpP{4.52\%})& 53.09\% (\relativeimpP{7.68\%}) & 47.39\%  (\relativeimpP{1.98\%})  \\
InfoVQA\textsubscript{val}~\cite{infographicvqa} & 32.63\% & 31.10\% (\relativeimpN{1.54\%})  & 32.42\% (\relativeimpN{0.21\%})& 32.20\%  (\relativeimpN{0.43\%})& 33.52\%  (\relativeimpP{0.89\%}) & 32.60\% (\relativeimpN{0.04\%})   \\
MMMU\textsubscript{val}~\cite{yue2024mmmu}    & 38.33\% & 39.11\% (\relativeimpP{0.78\%})  & 39.00\% (\relativeimpP{0.67\%}) & 39.89\% (\relativeimpP{1.56\%}) & 39.78\% (\relativeimpP{1.45\%}) & 39.56\%  (\relativeimpP{1.22\%})  \\
OCRBench~\cite{ocrbench}    & 36.30\% & 37.70\% (\relativeimpP{1.40\%})  & 37.10\% (\relativeimpP{0.80\%})& 40.00\% (\relativeimpP{3.70\%})& 38.60\% (\relativeimpP{2.30\%})& 36.50\%   (\relativeimpP{0.20\%}) \\
POPE~\cite{pope}       & 87.53\% & 87.74\% (\relativeimpP{0.21\%})  & 87.54\% (\relativeimpP{0.01\%})& 88.20\% (\relativeimpP{0.67\%}) & 86.81\% (\relativeimpN{0.72\%}) & 87.58\%  (\relativeimpP{0.04\%})  \\
QBench2\textsubscript{dev}~\cite{qbench} & 47.20\% & 52.40\% (\relativeimpP{5.20\%})  & 46.90\% (\relativeimpN{0.30\%})& 49.10\% (\relativeimpP{1.90\%}) & 48.90\%  (\relativeimpP{1.70\%})& 49.30\%  (\relativeimpP{2.10\%})  \\
RealworldQA~\cite{realworldqa}  & 57.25\% & 57.39\% (\relativeimpP{0.13\%})  & 59.35\% (\relativeimpP{2.09\%})& 57.52\% (\relativeimpP{0.26\%}) & 59.22\% (\relativeimpP{1.96\%}) & 57.91\%  (\relativeimpP{0.65\%})   \\
ScienceQA~\cite{scienceqa}   & 77.65\% & 78.31\% (\relativeimpP{0.66\%})  & 75.45\% (\relativeimpN{2.19\%})& 76.82\%  (\relativeimpN{0.83\%})& 77.60\% (\relativeimpN{0.05\%}) & 79.34\%  (\relativeimpP{1.70\%})  \\
SeedBench~\cite{seedbench}    & 69.23\% & 68.45\% (\relativeimpN{0.79\%})  & 68.26\% (\relativeimpN{0.98\%})& 69.54\% (\relativeimpP{0.30\%})& 68.78\%  (\relativeimpN{0.46\%}) & 68.95\% (\relativeimpN{0.28\%})   \\
TextVQA\textsubscript{val}~\cite{textvqa} & 56.29\% & 55.08\% (\relativeimpN{1.20\%})  & 56.44\% (\relativeimpP{0.15\%})& 60.97\% (\relativeimpP{4.68\%}) & 57.00\%  (\relativeimpP{0.71\%})& 56.12\%  (\relativeimpN{0.17\%}) \\
\bottomrule
\bottomrule
\end{tabular}
\vspace{-0.2cm}
\caption{\textbf{Main results.} We report the results obtained by training the LlaVA-Next model on the data curated with our proposed method (denoted as \textit{Our + [Method]}) and compare it with models trained on the data proposed in the respective publications and curated by their proprietary codes and GPT4 (denoted as \textit{[Method]}). In all cases, either original or our curated data is combined with the LlaVA-mix (LlaVA-1.5 fine-tuning data). All results are compared to LlaVA-mix-only training to obtain gains (green) and losses (red) in parenthesis.}
\label{tab:table1}
\end{table*}

% \begin{table}[ht]
% \begin{tabular}{M{2cm}M{2cm}M{2cm}cccccccccccccc}
% \toprule
% Scaling & Jacob-LLaVA1.5 (max \# conversations, 3x) + LLaVA Base & 
% Jacob-LLaVA1.5 (max images, 5x from coco captions) + LLaVA Base\\
% \midrule
% Ours \\
% 1$\times$        &           &                            \\
% 2$\times$    &           &                            \\
% 3$\times$   &           &                           \\
% \bottomrule
% \end{tabular}
% \caption{Ablation on scaling 1) number of conversations per image 2) number of images.}
% \label{tab:ablation_scaling_img_or_conversation}
% \end{table}

% \begin{table*}[ht]
% \begin{tabular}{cccccccM{1.5cm}M{1.5cm}M{1.5cm}M{1.5cm}ccccccccccccc}
% \toprule
% LLaVA  & Our LLaVA &  +more data sources & +more data sources + 2$\times$ Instruct. Data\\
% \midrule
% 52.36\% &   52.94\% (\relativeimpP{0.58\%})   &  53.68\% (\relativeimpP{1.32\%})  & 54.92\% (\relativeimpP{2.56\%})    \\

% \bottomrule
% \end{tabular}
% \caption{Ablation on scaling 
% %1) number of conversations per image 2) number of images.
% }
% \label{tab:ablation_scaling_img_or_conversation}
% \end{table*}
\begin{table*}[ht]
\scriptsize
\centering
\begin{tabular}{lcccc}
\toprule
Model & LLaVA & Our LLaVA  & + 14 Meta. Sources  & + 14 Meta. Sources + 2$\times$ Instruct. Data  \\
\midrule
AI2D~\cite{ai2d} & 62.82\% & 63.18\%  (\relativeimpP{0.36\%}) & 63.60\%  (\relativeimpP{0.78\%}) & 62.66\%  (\phantom{0}\relativeimpN{0.16\%}) \\
ChartQA~\cite{masry-etal-2022-chartqa} & 17.68\% & 19.44\%  (\relativeimpP{1.76\%}) & 19.64\%  (\relativeimpP{1.96\%}) & 30.56\%  (\relativeimpP{12.88\%}) \\
DocVQA\textsubscript{val}~\cite{docvqa} & 45.41\% & 45.43\%  (\relativeimpP{0.02\%}) & 47.78\%  (\relativeimpP{2.37\%}) & 49.37\%  (\phantom{0}\relativeimpP{3.96\%}) \\
InfoVQA\textsubscript{val}~\cite{infographicvqa} & 32.63\% & 31.10\%  (\relativeimpN{1.54\%}) & 33.02\%  (\relativeimpP{0.39\%}) & 34.17\%  (\phantom{0}\relativeimpP{1.54\%}) \\
MMMU\textsubscript{val}~\cite{yue2024mmmu} & 38.33\% & 39.11\%  (\relativeimpP{0.78\%}) & 39.11\%  (\relativeimpP{0.78\%}) & 39.11\%  (\phantom{0}\relativeimpP{0.78\%}) \\
OCRBench~\cite{ocrbench} & 36.30\% & 37.70\%  (\relativeimpP{1.40\%}) & 38.10\%  (\relativeimpP{1.80\%}) & 39.30\%  (\phantom{0}\relativeimpP{3.00\%}) \\
POPE~\cite{pope} & 87.53\% & 87.74\%  (\relativeimpP{0.21\%}) & 88.11\%  (\relativeimpP{0.58\%}) & 88.48\%  (\relativeimpP{0.94\%}) \\
QBench2\textsubscript{dev}~\cite{qbench} & 47.20\% & 52.40\%  (\relativeimpP{5.20\%}) & 45.00\%  (\relativeimpN{2.20\%}) & 49.50\%  (\phantom{0}\relativeimpP{2.30\%}) \\
RealworldQA~\cite{realworldqa} & 57.25\% & 57.39\%  (\relativeimpP{0.13\%}) & 58.95\%  (\relativeimpP{1.70\%}) & 59.35\%  (\phantom{0}\relativeimpP{2.09\%}) \\
ScienceQA~\cite{scienceqa} & 77.65\% & 78.31\%  (\relativeimpP{0.66\%}) & 78.24\%  (\relativeimpP{0.59\%}) & 78.57\%  (\phantom{0}\relativeimpP{0.92\%}) \\
SeedBench~\cite{seedbench} & 69.23\% & 68.45\%  (\relativeimpN{0.79\%}) & 68.78\%  (\relativeimpN{0.46\%}) & 68.65\%  (\phantom{0}\relativeimpN{0.58\%}) \\
TextVQA\textsubscript{val}~\cite{textvqa} & 56.29\% & 55.08\%  (\relativeimpN{1.20\%}) & 57.77\%  (\relativeimpP{1.48\%}) & 59.32\%  (\phantom{0}\relativeimpP{3.04\%}) \\
\midrule
Average & 52.36\% & 52.94\%  (\relativeimpP{0.58\%}) & 53.12\%  (\relativeimpP{0.81\%}) & 54.92\%  (\relativeimpP{2.56\%}) \\
\bottomrule
\end{tabular}
\vspace{-0.2cm}
\caption{\textbf{Scalability of our approach} We report results with additional data sources and instruction-following data. We compare the baseline LLaVA-mix training (\textit{LLaVA}), our LLaVA-mix with LLaVA-Instruct-150K replaced (\textit{Our LLaVA}) together with two variants further scaled by our approach. 
The \textit{+14 Meta. Sources} variant uses 14 additional metadata sources (detailed in Supplementary). The \textit{+14 Meta. Sources + 2$\times$ Instruct. Data } variant further scales 2$\times$ instruction data.    }
\vspace{-0.1cm}
\label{tab:actual_table}
\end{table*}

\begin{table}[ht]
\scriptsize
\begin{tabular}{M{1.5cm}M{1.5cm}M{1.5cm}M{1.5cm}cccccccccccc}
\toprule
 LLaVA & Gemma2 27B~\cite{gemma2} & LLaMA~3.1 70B~\cite{llama3_1}\\
\midrule
52.36\% &   52.94\% (\relativeimpP{0.58\%})   &  52.17\% (\relativeimpN{0.19\%})   \\
\bottomrule
\end{tabular}
\vspace{-0.2cm}
\caption{\textbf{Ablation on the choice of LLM.} We report results by employing an additional LLM, as compared to the LLM used for the main results, highlighting the generalizability of our approach.}
\vspace{-0.1cm}
\label{tab:ablation_llm_choice}
\end{table}

% \input{tables/mytable2}
% \input{tables/ablation1}
% \input{tables/ablation2}

% \wei{TODO R3 thinks this introduction paragraph is too long.}

\subsection{Implementation Details}

\myparagraph{Multimodal Models.}
We conducted experiments to reproduce the metadata-to-visual-instructions conversion proposed in several recent multimodal models. Notably, we replicated LLaVA-1.5 \cite{llava}, the seminal work that introduced visual instruction tuning and the LLaVA-Instruct-150K dataset, which was originally generated by prompting GPT-4. Additionally, we successfully reproduced instruction sets from other works that were also generated using GPT-4, such as PF-1M \cite{pf1m} and LLaVAR \cite{llavar}. More results on additional instruction sets produced by GPT-4 are available in the Supplementary. Importantly, all these original instruction sets were generated using proprietary, unreleased code, with only brief descriptions of their dataset construction provided in the original publications. Here, we demonstrate that our approach—soon to be open-sourced—can effectively and reliably reproduce these high-quality instruction sets, making them more accessible for further research and development.

\myparagraph{Reproduction of Instruction Sets.} 
We apply our approach to reproduce (and potentially improve) instruction sets originally generated using proprietary GPT-4-based pipelines, including: (1) LLaVA-Instruct-150K \cite{llava} (2) PF-1M \cite{pf1m} (3) LLaVAR \cite{llavar}. For each dataset, we generate the same number of instructions for the same images as in the original sets and compare fine-tuning performance using (1) LLaVA-mix + the original dataset, and (2) LLaVA-mix + our reconstructed dataset. See below for reproduction details.

For \textbf{LLaVA-Instruct-150K}~\cite{llava}, we used the three original prompts from \cite{llava} for generating conversation, complex reasoning, and detailed captioning tasks, adding an output formatting instruction for consistency. The complete prompts used in our reproduction are provided in the Supplementary. For \textbf{PF-1M}~\cite{pf1m}, due to computational constraints, we applied our approach to a random subset of 19 out of the 37 datasets used in PF-1M (the full list is provided in the Supplementary). We also simplified their instruction formation to a single unified prompt, which we used with our staged generation approach. For \textbf{LLaVAR}~\cite{llavar}, we utilized the original prompts but replaced their OCR system (PaddleOCR) with the open-source EasyOCR package for text extraction.
Further details on data sources and implementation are provided in the Supplementary. 
%\wei{The details of data sources can be found in Supplementary. } % How about introduction of the evaluation benchmark datasets? 

\myparagraph{Evaluation of Instruction Sets.} 
We evaluate the efficacy of different instruction sets by fine-tuning LLaVA-Next \cite{llava-next}, which consists of an image encoder (CLIP \cite{clip}), an LLM decoder (Phi3-instruct-128k \cite{abdin2024phi}), and a projector (2-layer MLP), pre-trained following the LLaVA-1.5 methodology using a pre-training set composed of captioning instructions sampled from caption datasets and mechanically transformed. 
Our baseline model is trained on LLaVA-mix \cite{llava}, the original fine-tuning instruction set of LLaVA-1.5. Notably, the only portion of LLaVA-mix generated via an LLM (GPT-4) is the LLaVA-Instruct-150K subset.
We evaluate all fine-tuned models on 12 popular benchmarks.

\subsection{Reproducing and Improving Instruction Sets}\label{sec:exp_reproducing}
As shown in Table \ref{tab:table1}, our method successfully reproduces and even improves existing instruction sets. We observe consistent overall improvements and notable gains in specific ones across the 12 evaluation benchmarks. despite using lighter, open-source models rather than the proprietary GPT-4 models used in the original works. This highlights the broad applicability of our approach, demonstrating that high-quality instruction sets can be generated effectively without requiring expensive, proprietary models. Our method enables reproducibility while offering a scalable and accessible alternative to prior instruction generation techniques.

\subsection{Scaling for Enhanced Performance}\label{sec:exp_scaling}
To further demonstrate generality and scalability, we conduct experiments that expand metadata sources and increase the volume of generated instructions. The results, presented in Table \ref{tab:actual_table}, compare (1) Baseline LLaVA-mix training (2) Our LLaVA-mix variant (with LLaVA-Instruct-150K replaced) and (3) Two further scaled versions of our approach. 

For scaling meta sources, the \textit{+14 Meta. Sources} variant retains the same LLaVA-mix images but integrates 14 additional metadata sources (detailed in the Supplementary). This expansion significantly enhances performance across most benchmarks. The only exception is QBench which focuses on low-level vision tasks that might be somewhat biased away with the enhancement of sources and hence the richness and diversity of other instructions when done without adding additional images. 

For scaling instruction quantity, we further assess the impact of instruction quantity scaling by doubling the number of generated instructions while keeping the expanded metadata set. This results in an additional performance boost, with an average improvement of 5$\times$ across the 12 benchmarks. These results confirm that both metadata diversity and instruction volume contribute to better performance, demonstrating the robust scalability of our approach.

%%%%%%%%%%%%%% Ablations 
\subsection{Ablations}\label{sec:ablation}
This section analyzes the impact of key design choices in our approach, including the choice of LLM, the effectiveness of the ASCII tree structure for bounding box conversion, and the quality filtering mechanism. We evaluate these components through targeted experiments and quantitative assessments. Additional examples of automatically generated instructions for various image types are provided in the Supplementary Material.
\subsubsection{Choice of LLM}
To assess the robustness of our method with different LLMs, we reproduce the LLaVA-Instruct-150K experiment using LLaMA 3.1 70B \cite{llama3_1} instead of Gemma2 27B \cite{gemma2}. The average performance across 12 benchmarks is reported in Table \ref{tab:ablation_llm_choice}, with full benchmark-specific results provided in the Supplementary. Our findings indicate that both models enable us to closely reproduce the effectiveness of the original LLaVA-Instruct-150K while replacing GPT-4 with significantly smaller open models. Interestingly, despite its smaller size, Gemma2 27B outperforms LLaMA 3.1 70B in generating effective visual instructions. This suggests that model-specific properties, beyond just parameter count, play a crucial role in instruction generation quality.

\subsubsection{Effectiveness of the ASCII Tree Approach}
% \wei{R3 mentions that this ablation is not thorough enough. The effectivness of SAM2 and Depth Anything v2 and the impact of counting logic thresholds remain unknown. }
To evaluate the impact of our hierarchical ASCII tree structure for bounding box conversion compared to the standard concatenation-based approach, we conducted a detailed study on the LVIS dataset, a strong benchmark for bounding box-based tasks. We manually assessed the factual correctness of each conversation turn, considering responses valid if they aligned with either ground-truth image content or raw metadata annotations (e.g., object counts were deemed correct if they matched either the visual content or LVIS annotations).
\begin{table}[t!]
\scriptsize
\begin{tabular}{M{2cm}M{1.5cm}M{1.5cm}M{1.5cm}M{1.5cm}}
\toprule
Method & Total Turns & Incorrect Turns & Error Rate \\
\midrule
Normal Concatenation & 173 & 32 & 18.5\% \\
\midrule
Tree Structure & 194 & 19 & 9.8\% \\
\bottomrule
\end{tabular}
\vspace{-0.2cm}
\caption{\textbf{Bounding box conversion approach.} Comparison of bounding box conversion approaches on LVIS v2 dataset. The tree structure approach shows substantially lower error rates while enabling more comprehensive conversations.}
\vspace{-0.1cm}
\label{tab:tree_ablation}
\end{table}
\begin{table}[t!]
\scriptsize
\begin{tabular}{M{2cm}M{1.5cm}M{1.5cm}M{1.5cm}M{1.5cm}}
\toprule
Metric & Value & Rate (\%) \\
\midrule
Precision & 0.83 & 83.3\% \\
True Negatives & 22 & 73.3\% \\
False Negatives & 2 & 6.7\% \\
True Positives & 5 & 16.7\% \\
False Positives & 1 & 3.3\% \\
\bottomrule
\end{tabular}
\vspace{-0.2cm}
\caption{Evaluation of quality filtering effectiveness on third conversation turns across 30 images.}
\label{tab:filtering_ablation}
\end{table}
As shown in Table \ref{tab:tree_ablation}, our tree-based approach significantly improves both conversation quality and accuracy. The standard concatenation method resulted in 173 conversation turns, with 32 incorrect responses (18.5\% error rate), whereas our hierarchical tree approach yielded 194 turns with only 19 incorrect responses (9.8\% error rate). This represents a relative error reduction of 47\%, demonstrating the effectiveness of our approach in structuring complex spatial information.
Several factors contribute to this improvement:
(1) Enhanced preservation of spatial relationships between objects
(2) More intuitive organization of object hierarchies
(3) Clearer representation of part-whole relationships
(4) Improved handling of overlapping and nested objects.
These findings validate our hierarchical tree structure as a superior method for converting spatial data into natural language descriptions.

\subsection{Effectiveness of Quality Filtering}
To assess the impact of our quality filtering mechanism, we analyzed the third conversation turn generated for each image in a sample of 30 images. Specifically, we evaluated whether the filtering process successfully identified and removed low-quality or irrelevant conversation turns.
As presented in Table \ref{tab:filtering_ablation}, our filtering approach achieved a precision of 0.83, indicating its effectiveness in detecting problematic conversation turns while preserving high-quality responses. This demonstrates that our filtering method is a reliable component for maintaining instruction relevance and coherence in generated conversations.

\section{Conclusion}
% We have proposed a complete, open-source solution \method for generating visual instruction tuning data that addresses the key challenges of scalability, diversity, repeatability, and quality control, enabling rapid development and testing of synthetic data generation for visual instruction tuning. Importantly, we have shown that it is possible to obtain this high-quality data using open models, without resorting to expensive closed-source models such as GPT4. Through extensive experimentation, our system was able to demonstrate compelling generalization to different previously proposed instruction tuning datasets, as well as a strong ability to scale with additional metadata sources and additional instructions produced by our approach. We have carefully analyzed the different components of the approach to demonstrate their choices and effectiveness. Some interesting \textit{future work} that can stem from our approach includes its iterative application on its own outputs combining them with additional conversational instruction data sources; extending it with chain-of-thought and other forms of stepwise reasoning support; combining it with data-mix optimization techniques for adaptively guiding the produced instruction curriculum.

We introduced \method, an open-source framework for generating high-quality visual instruction tuning data with scalability, diversity, and quality control. Our approach eliminates the need for expensive proprietary models like GPT-4, demonstrating strong generalization to existing instruction datasets and effective scalability with additional metadata and instructions. Through extensive analysis, we validated the effectiveness of each component. Future directions include iterative self-improvement, integration with stepwise reasoning, and data-mix optimization for adaptive instruction generation.

% \subsection{Key Takeaways}

% \begin{itemize}
%     \item \textbf{Unified, Open-Source Platform:} Enables rapid development and testing of synthetic data generation.
%     \item \textbf{Reduced Dependence on Closed-Source Models:} Demonstrates that high-quality data can be generated without GPT-4.
%     \item \textbf{Advancement in Visual Instruction Tuning:} Supports scalability, generalizability, and repeatability.
% \end{itemize}

% \input{sec/0_abstract}    
% \input{sec/1_intro}
% \input{sec/2_formatting}
% \input{sec/3_finalcopy}

\appendix
\section*{Supplementary}

For further insights into our approach \method, we 
% provide source code (Sec.~\ref{sec:source_code}), 
give an overview of data source including dataset descriptions, statistics, and data efficiency analysis (Sec.~\ref{sec:dataset_details}), and provide more implementation details on hardware setup and configurations. 

As additional results, we provide analysis of replacing GPT-4V-generated data in the mPLUG-DocOwl DocReasoning dataset with our \method in Sec.~\ref{sec:docowl_dataset_replacement}. We include a study on the contribution of different components in \method in Sec.~\ref{ablation_effectiveness_components}. We include experiments demonstrating Instructify’s dataset size scaling properties in Sec.~\ref{sec:scaling_dataset_size}. 

We also provide qualitative examples of reproduced conversations (Sec.~\ref{sec:reproduction_conversation_gallery}) and illustration of our hierarchical ASCII Tree representation (Sec.~\ref{sec:ascii_tree}). In the end, we give a summary of the input instructions used to generate instruction-tuning conversations across different datasets (Sec.~\ref{sec:prompts}).

\section{Dataset Details}\label{sec:dataset_details}
This section provides an overview of the datasets used to construct instruction-tuning conversations.
% We start with providing dataset descriptions in Section~\ref{subsec:dataset_descriptions}, then list details about data efficiency in Section~\ref{subsec:data_efficiency}.
\subsection{Dataset Descriptions}\label{subsec:dataset_descriptions}
The datasets utilized in this work are summarized in Table~\ref{table:DatasetGroup}, detailing the number of images, captions, bounding boxes, and QA pairs used for training. The total available data exceeds the subset selected for training.

\paragraph{LLaVA Reproduction:}
The LLaVA Reproduction dataset is derived entirely from COCO Captions \cite{cococaption}, incorporating both captions and bounding boxes.

\paragraph{Polite Flamingo~\cite{pf1m}:}
Our reproduction of the Polite Flamingo dataset includes most of the original sources. A complete list of input datasets is provided in Table~\ref{table:DatasetGroup}, many of which overlap with COCO and Visual Genome images.

\paragraph{LLaVAR~\cite{llavar}:}
LLaVAR integrates captions from a filtered subset of LAION-5B \cite{laion-5b} as provided by the original work \cite{llavar}. Additionally, we generate OCR annotations using EasyOCR \cite{easyocr}, structuring them as bounding boxes. Each image receives a single generated conversation. The full dataset contains 19,800 images, with 13,104 sampled for training (as shown in Table~\ref{table:DatasetGroup}).

\paragraph{Mix Dataset:}
The Mix dataset combines all data sources used in LLaVA \cite{llava} training, incorporating 14 additional datasets beyond the original COCO Captions. The full merged dataset contains 506,110 images, from which conversations were sampled for training: (1) \textit{1$\times$ Mix}: 160,643 sampled conversations (2) \textit{2$\times$ Mix}: 321,286 sampled conversations. 

For the +14 datasets, 2x Mix, additional sources such as ST-VQA, Chart-QA, and DocVQA were included in small proportions, making up only 0.5\% of the training data.

\begin{table*}[h]
\centering
\small
\caption{\textbf{Data Efficiency Results.} Feature Configuration vs. Performance (producing 500 conversations)}
\label{table:EfficiencyResults}
\resizebox{\textwidth}{!}{%
\begin{tabular}{l|ccc|cc}
\toprule
% \hline
\textbf{Variants} & \textbf{Filtering} & \textbf{BBox Conversion} & \textbf{Reduction} & \textbf{Time (s)} & \textbf{Throughput (conv/hour)} \\
% \hline
\midrule
Direct Generation &  &  &  & 50 & 36,000 \\
+Filtering & \checkmark &  &  & 60 & 30,000 \\
+BBox &  & \checkmark &  & 304 & 5,900 (imgs/hour)\\
+Reduction &  &  & \checkmark & 51 & 35,300 \\
BBox SAM2/DepthV2 Only &  & \checkmark &  & 396 & 4,500 (imgs/hour) \\
Full Processing & \checkmark & \checkmark & \checkmark  & 191 & 9,400 \\
% \hline
\bottomrule
\end{tabular}%
}
\end{table*}
\subsection{Data Efficiency}
\label{subsec:data_efficiency}
We assessed the efficiency of our data pipeline using two NVIDIA A100 80GB GPUs, analyzing the impact of filtering, bounding box (BBox) conversion, and data reduction. Performance results under various configurations are summarized in Table~\ref{table:EfficiencyResults}.

With all optimizations enabled—including filtering, BBox conversion, and reduction—the pipeline generates 9,400 conversations per hour when processing 500 images. For a smaller batch of 100 images, it produces 502 filtered conversations in 216 seconds. The most computationally intensive step is BBox conversion, with SAM2 and Depth Anything V2 being the primary contributors to processing time.

%%%%%%%%%%%%%%%%%%%%%%%%%%%%%%% implementation details 
\section{Implementation Details}\label{sec:implementation_details}
Our system is implemented in Python with a strong emphasis on scalability and reproducibility. It supports any models compatible with \texttt{vLLM}, a widely used backend for efficient LLM inference. Our experiments were conducted on NVIDIA A100 GPUs within a cluster environment, enabling horizontal scaling for high-speed, large-scale data generation. Detailed efficiency metrics are provided in the Supplementary.
Unless otherwise specified, Gemma2-27B \cite{gemma2} was used as the primary LLM for instruction data generation. All models were evaluated on 12 popular LMM benchmarks using the LMMs-Eval harness \cite{zhang2024lmmsevalrealitycheckevaluation}, ensuring a standardized and comprehensive assessment.
The system's modular design allows researchers to modify and extend individual components while maintaining pipeline integrity. Key parameters, including prompt distributions, quality thresholds, and processing configurations, are fully configurable via external settings. Our code is available in the Supplementary and will be released publicly upon acceptance.

%%%%%%%%%%%%%%%%%%%%%%%%%%%%%%% dodcowl dataset reproduction 
\begin{table}[h]
\scriptsize
\begin{tabular}{l|c|cc}
\toprule
    Benchmark         & LLaVA & +DocReason (\%Rel) & +Our DocReason (\%Rel) \\
\midrule
AI2D                 & 0.6282 & 0.6418 (\relativeimpP{1.36\%}) & 0.6324 (\relativeimpP{0.42\%}) \\
ChartQA              & 0.1768 & 0.2336 (\relativeimpP{5.68\%}) & 0.3820 (\relativeimpP{20.52\%}) \\
DocVQA\textsubscript{val} & 0.4541 & 0.5696 (\relativeimpP{11.55\%}) & 0.5564 (\relativeimpP{10.23\%}) \\
InfoVQA\textsubscript{val} & 0.3263 & 0.3488 (\relativeimpP{2.25\%}) & 0.3681 (\relativeimpP{4.18\%}) \\
MMMU\textsubscript{val}    & 0.3833 & 0.3967 (\relativeimpP{1.33\%}) & 0.4022 (\relativeimpP{1.89\%}) \\
OCRBench             & 0.3630 & 0.4150 (\relativeimpP{5.20\%}) & 0.4180 (\relativeimpP{5.50\%}) \\
POPE                 & 0.8753 & 0.8732 (\relativeimpN{0.21\%}) & 0.8679 (\relativeimpN{0.74\%}) \\
QBench2\textsubscript{dev} & 0.4720 & 0.4820 (\relativeimpP{1.00\%}) & 0.4660 (\relativeimpN{0.60\%}) \\
RealworldQA          & 0.5725 & 0.5882 (\relativeimpP{1.57\%}) & 0.5712 (\relativeimpN{0.13\%}) \\
ScienceQA            & 0.7765 & 0.7838 (\relativeimpP{0.73\%}) & 0.7861 (\relativeimpP{0.97\%}) \\
SeedBench            & 0.6923 & 0.6879 (\relativeimpN{0.44\%}) & 0.6838 (\relativeimpN{0.85\%}) \\
TextVQA\textsubscript{val} & 0.5629 & 0.5653 (\relativeimpP{0.24\%}) & 0.5781 (\relativeimpP{1.53\%}) \\
\bottomrule
\end{tabular}
\caption{\textbf{Results on DocOwl Dataset Reproduction.} Actual and relative performances differences for replacing mPlug DocReasoning\cite{mplug_docowl}. Gains (\relativeimpP{}) and losses (\relativeimpN{}) are shown in parentheses.}
\label{tab:docowl_comparison}
\end{table}

\section{DocOwl Dataset Reproduction}~\label{sec:docowl_dataset_replacement}
We further evaluate the impact of replacing GPT-4V-generated data from the mPLUG-DocOwl DocReasoning dataset \cite{mplug_docowl}. We train a model using LLaVA data augmented with the original DocReasoning dataset.

For our reproduction of GPT-4V-generated data, we leverage the same OCR collection as our LLaVAR system and adapt mPLUG’s instruction format to guide Gemma 2 27B in generating conversations solely from metadata. As shown in Table~\ref{tab:docowl_comparison}, both models demonstrate substantial improvements in document reasoning across multiple benchmarks. However, our reproduced dataset yields even greater gains. While the addition of the original DocReasoning dataset results in an average relative improvement of 2.52\%, our reproduced dataset achieves a higher average relative improvement of 3.58\%.

%%%%%%%%%%%%%%%%%%%%%%%%%%%%%%% ablation on effectiveness of components
\begin{table}
% \vspace{-0.5cm}
    \centering
    % \scriptsize
    \begin{tabular}{lcr}
        \toprule
        \textbf{Configuration} & \textbf{Avg. Errors} & \textbf{$\boldsymbol{\Delta}$} \\
        \midrule
        Base System & 2.88 & -- \\
        \midrule
        w/o Tree Structure & 3.47 & +0.59 \\
        w/o Box Labels & 3.04 & +0.16 \\
        w/o Depth Info & 2.92 & +0.04 \\
        w/o Object Counting & 2.47 & - 0.41 \\
        \bottomrule
    \end{tabular}
    % \vspace{-0.2cm}
    \caption{Ablating different components of Instructify.  
    }
    \label{tab:effectiveness_of_comp}
    % \vspace{-0.4cm}
\end{table}
\section{Ablation on Effectiveness of Components}\label{ablation_effectiveness_components}
Table~\ref{tab:effectiveness_of_comp} presents an ablation study assessing the impact of different components in our Instructify system. We evaluate the factual correctness of generated conversations on 100 randomly sampled COCO images, using GPT-4o for evaluation.

Our findings indicate that every component contributes to data quality. Notably, the most significant performance drop occurs when removing the hierarchical tree structure, leading to an increase in errors by up to 20.5\%.

%%%%%%%%%%%%%%%%%%%%%%%%%%%%%%% scaling of dataset size 
\begin{table}
\centering
% \scriptsize
% \resizebox{0.3\textwidth}{!}
{\begin{tabular}{l|c|c|c}
\toprule
Metric & Base & 150K & 825K \\
\midrule
chartqa & 0.1632 & 0.1844 \relativeimpnP{2.12} & 0.1968 \relativeimpnP{3.36} \\
docvqa\textsubscript{val} & 0.3788 & 0.4463 \relativeimpnP{6.75} & 0.4836 \relativeimpnP{10.48} \\
gqa & 0.6192 & 0.6209 \relativeimpnP{0.17} & 0.6348 \relativeimpnP{1.56} \\
infovqa\textsubscript{val} & 0.2942 & 0.3127 \relativeimpnP{1.85} & 0.3256 \relativeimpnP{3.14} \\
ocrbench & 0.3290 & 0.3730 \relativeimpnP{4.40} & 0.3980 \relativeimpnP{6.90} \\
textvqa\textsubscript{val} & 0.5106 & 0.5407 \relativeimpnP{3.01} & 0.5856 \relativeimpnP{7.50} \\
\bottomrule
\end{tabular}}
% \vspace{-0.2cm}
\caption{{Scaling Conversation Sampling}.}
\label{tab:scaling_exp}
% \vspace{-0.5cm}
\end{table}
\section{Scaling of Dataset Size}\label{sec:scaling_dataset_size}
Table~\ref{tab:scaling_exp} presents experiments on the impact of scaling dataset size for training. Our results show that Instructify exhibits strong scaling properties, achieving up to 10.48\% improvement with increased instruction data.

%%%%%%%%%%%%%%%%%%%%%%%%%%%%%%% reproduction conversation gallery 
\section{Reproduction Conversation Gallery}\label{sec:reproduction_conversation_gallery}
We provide qualitative samples of our Reproduction Conversation Gallery in the file \texttt{reproduction\_examples.zip}. The zip file includes the following:
(1) \textit{LLaVA Reproduction Conversation Gallery} (2) \textit{LLaVA Reproduction +14 Datasets Mix Conversation Gallery} (3) \textit{LLaVAR Reproduction Mix Conversation Gallery} (4) \textit{DocOwl DocReason Reproduction Mix Conversation Gallery}. 

These samples illustrate the effectiveness of our approach in reproducing and enhancing instruction tuning datasets.

% \wei{Reproduction conversation gallery: LLaVA Reproduction, DocOwl DocReason Reproduction Mix, LLaVAR Reproduction Mix, LLaVA Reproduction +14 Datasets Mix,     in the html file,  "replacement" refers to reproduction }

%%%%%%%%%%%%%%%%%%%%%%%%%%%%%%%  ASCII tree examples 
\section{ASCII Tree Examples}\label{sec:ascii_tree}
\begin{figure*}[ht]
% \vspace{-0.3cm}
\centering
\includegraphics[width=0.9\linewidth]{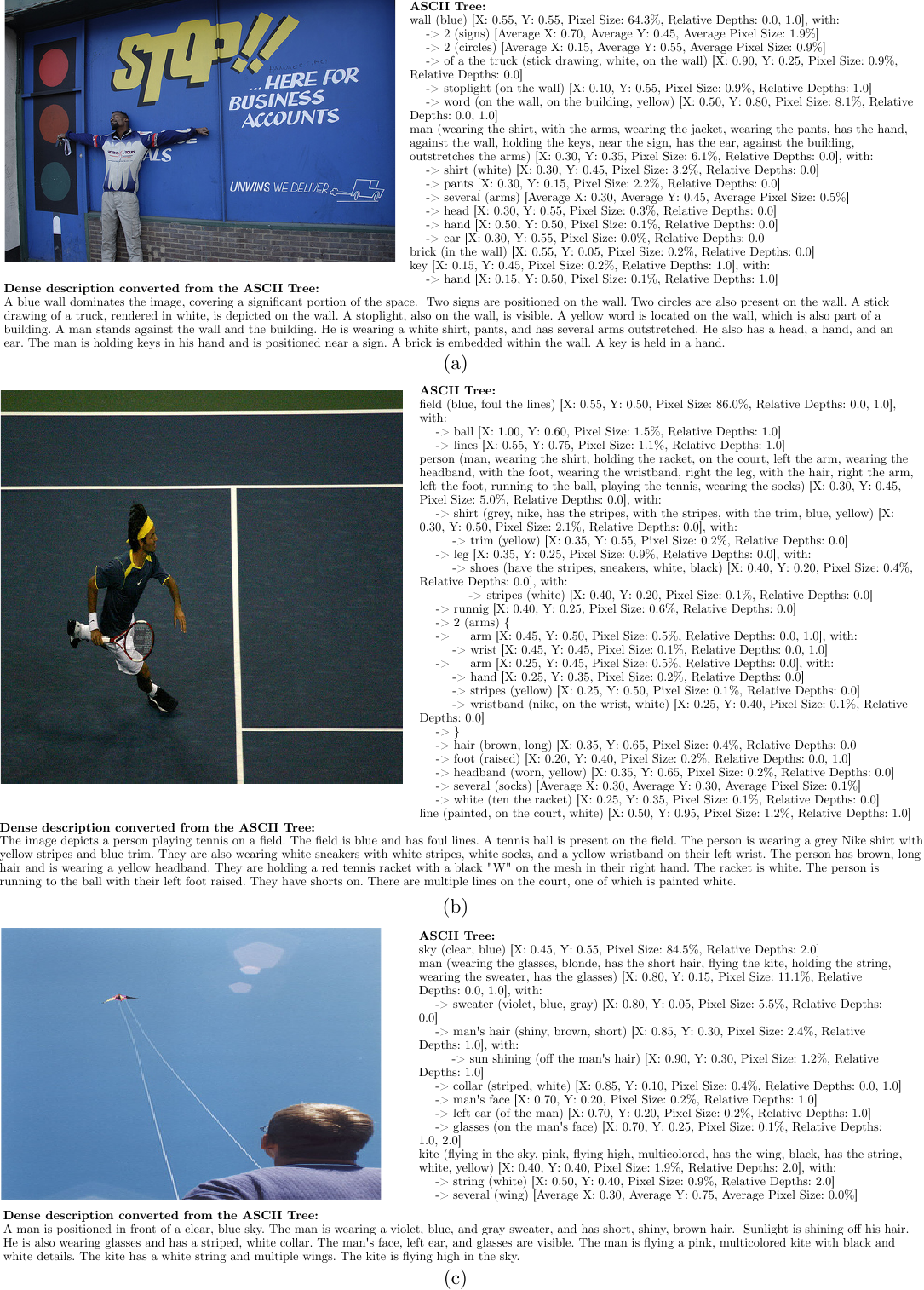}
\caption{
\label{fig:supp_ascii_tree_examples}
\textbf{ASCII tree example.} Our approach automatically collects metadata from different available for the image. The ASCII Tree (top right) is a hierarchical textual data structure that includes the attributes, positions, sizes and depth information of objects in the image. The ASCII tree is further converted by an LLM into a high-quality dense description (bottom) consisting of multiple factual statements. 
}
\end{figure*}

% More examples of ASCII Tree. Fig.~\ref{fig:supp_ascii_tree_examples}
Figure~\ref{fig:supp_ascii_tree_examples} presents three examples of our ASCII Tree representation. For each image, our method automatically extracts metadata—such as object attributes, positions, sizes, and depth information—and organizes it into a hierarchical textual structure. This structured data is then processed by a large language model (LLM) to generate detailed, high-quality descriptions composed of multiple factual statements, effectively summarizing the image content.

%%%%%%%%%%%%%%%%%%%%%%%%%%%%%%% prompts 
\section{Prompts}\label{sec:prompts}
\begin{figure*}[ht]
% \vspace{-0.3cm}
\centering
\includegraphics[width=0.95\linewidth]{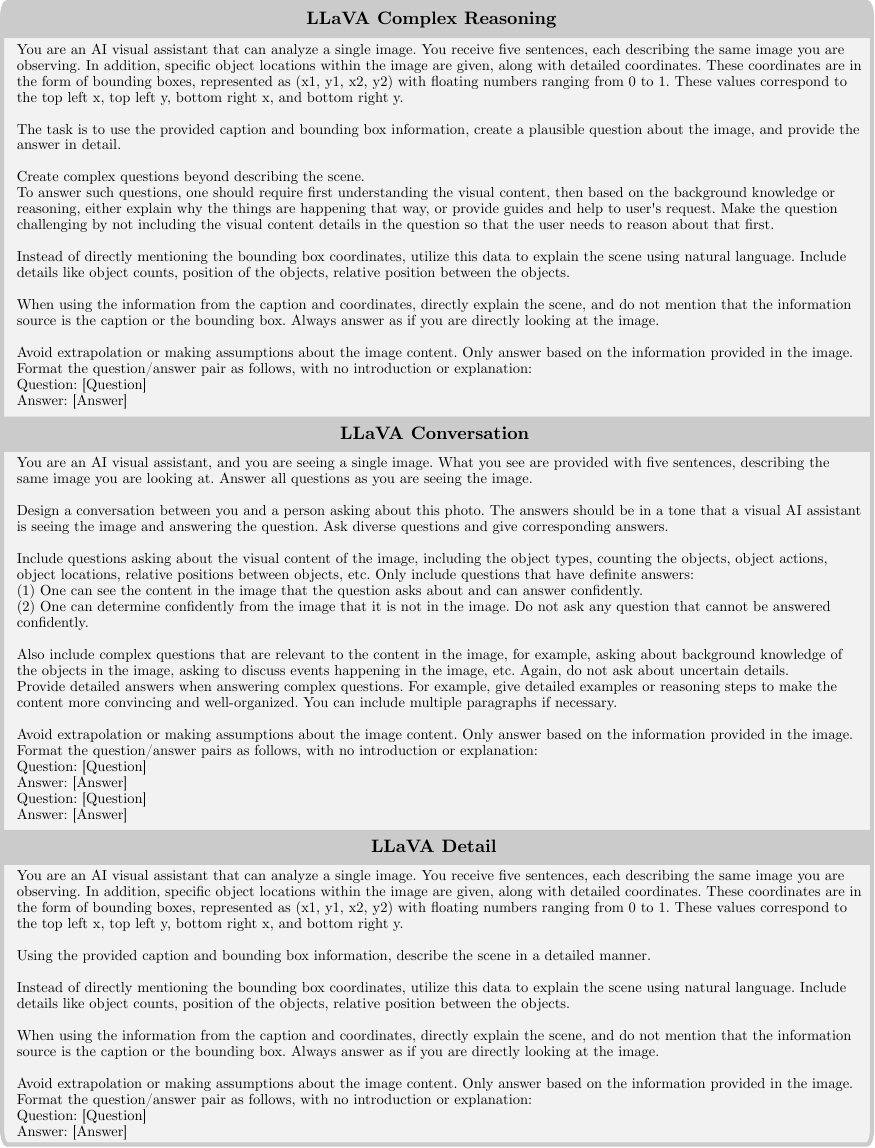}
\caption{
\label{fig:prompt_llava}
\textbf{LLaVA prompt.} 
}
\end{figure*}

\begin{figure*}[ht]
% \vspace{-0.3cm}
\centering
\includegraphics[width=0.95\linewidth]{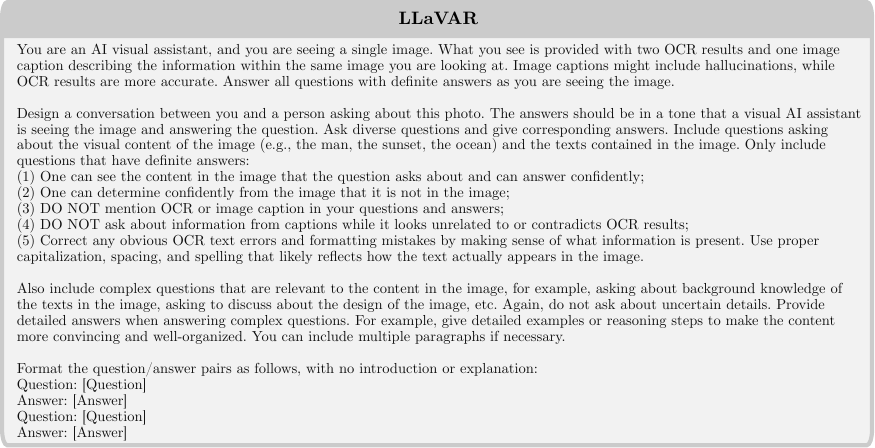}
\caption{
\label{fig:prompt_LLaVAR}
\textbf{LLaVAR prompt.}}
\end{figure*}

\begin{figure*}[ht]
% \vspace{-0.3cm}
\centering
\includegraphics[width=0.95\linewidth]{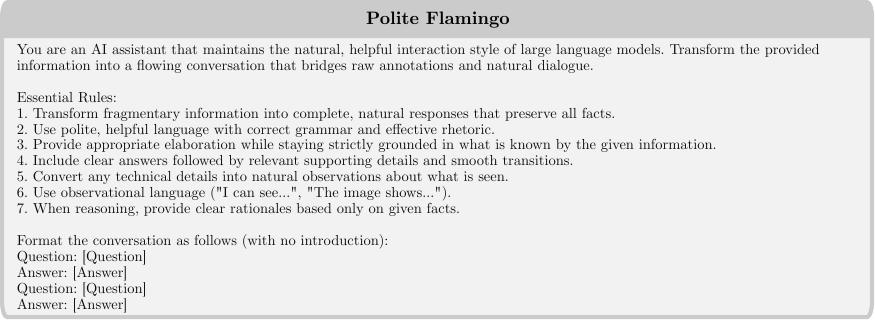}
\caption{
\label{fig:prompt_PF}
\textbf{Polite Flamingo prompt.} 
}
\end{figure*}

\begin{figure*}[ht]
% \vspace{-0.3cm}
\centering
\includegraphics[width=0.95\linewidth]{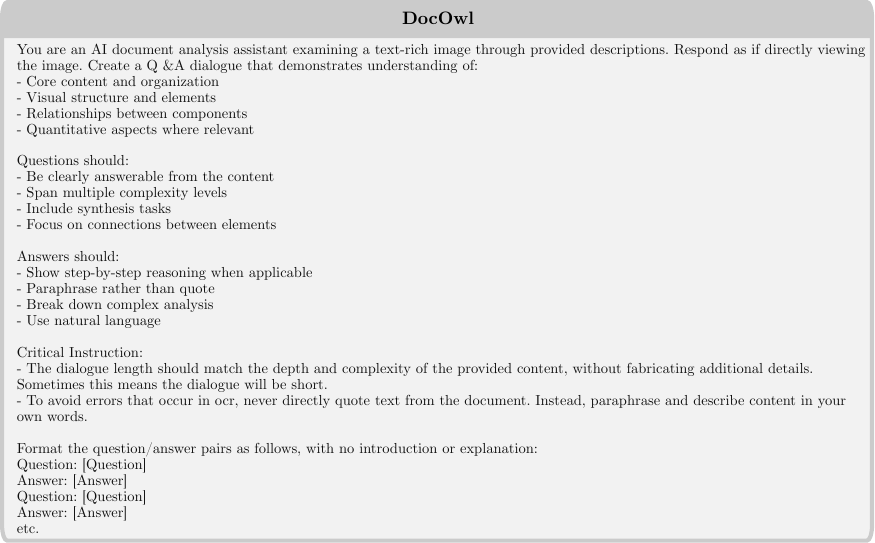}
\caption{
\label{fig:prompt_docowl}
\textbf{DocOwl prompt.}
}
\end{figure*}

The prompts used in this study define the interaction style and reasoning approach of the generated conversations. Below, we provide an overview of the input instructions for each task, adapted from the original papers: 
(1) LLaVA Prompt – Figure~\ref{fig:prompt_llava}
(2) LLaVAR Prompt – Figure~\ref{fig:prompt_LLaVAR}
(3) Polite Flamingo Prompt – Figure~\ref{fig:prompt_PF}
(4) DocOwl Prompt – Figure~\ref{fig:prompt_docowl}. 

These prompts guide the structured generation of conversations, ensuring diverse and contextually appropriate responses.

% prompt LLaVA Fig.~\ref{fig:prompt_llava}\\
% prompt LLaVAR Fig.~\ref{fig:prompt_LLaVAR}\\
% prompt Polite Flamingo Fig.~\ref{fig:prompt_PF} \\
% prompt DocOwl Fig.~\ref{fig:prompt_docowl}\\
% prompt Shikra Fig.~\ref{fig:prompt_shikra}\\ 

% \include{figure_tex/supp_prompt_shikra}

\begin{table*}[htbp]
\centering
\scriptsize
\caption{\textbf{Dataset Contributions.} We list the dataset contribution to captions, bounding boxes, and QA Tasks. Here \textbf{mix} is the mixture of coco captions combined with the 14 datasets.}
\begin{tabular}{lccccccccc}
% \hline

%%%%%%%%%%%%%%%%%%%%%% LLaVA-1.5
\toprule
\multicolumn{5}{c}{\textbf{LLaVA~\cite{llava}}} \\
\midrule
\textbf{Dataset Group} & \textbf{\# Images} & \textbf{\# Captions} & \textbf{\# Bounding Boxes} & \textbf{\# QA} \\
\midrule

COCO Captions\cite{cococaption} & 117,702 & 588,827 & 856,988 & 0\\
\midrule
Total & 117,702 & 588,827 & 856,988 & 0\\
\bottomrule

%%%%%%%%%%%%%%%%%%%%%% Polite Flamingo
\hline
\hline
\toprule
\multicolumn{5}{c}{\textbf{Polite Flamingo~\cite{pf1m}}} \\
\midrule
% \textbf{Dataset Group} & \textbf{\# Images} & \textbf{\# Captions} & \textbf{\# Bounding Boxes} & \textbf{\# QA} \\
% \midrule
% \hline
% \hline

A-OKVQA\cite{a-okvqa} & 10,407 & 0 & 0 & 10,407\\
COCO Captions\cite{cococaption} & 79,866 & 399,513 & 583,502 & 0\\
DIOR RSVG\cite{rsvg} & 8,866 & 0 & 39,644 & 0\\
Flickr 30K\cite{flickr30k} & 30,501 & 152,504 & 0 & 0\\
GQA\cite{gqa} & 80,356 & 0 & 6,201,642 & 900,764\\
TGRS-HRRSD\cite{TGRS-HRRSD} & 13,458 & 13,458 & 47,162 & 0\\
image2paragraph\cite{image2paragraph} & 14,194 & 14,198 & 0 & 0\\
Image Editing Request\cite{imageediting} & 838 & 838 & 0 & 0\\
LEVIR\cite{levir}\_cc & 9,870 & 49,350 & 0 & 0\\
OCR-VQA\cite{ocrvqa} & 95,719 & 282,928 & 0 & 464,978\\
refCOCOg\cite{refcoco} & 20,975 & 0 & 77,000 & 0\\
RemoteCLIP Ret-3\cite{remoteclip} & 13,408 & 81,888 & 0 & 0\\
Science QA\cite{scienceqa} & 5,876 & 5,876 & 0 & 2,591,240\\
TextCaps\cite{textcaps} & 20,968 & 1,152,690 & 0 & 0\\
TextVQA\cite{textvqa} & 20,968 & 20,968 & 0 & 33,054\\
Visual7w\cite{visual7w} & 5,628 & 0 & 0 & 27,222\\
VQA-E\cite{vqae} & 66,931 & 0 & 0 & 151,092\\
VQA V2\cite{vqav2} & 79,866 & 0 & 0 & 428,183\\

\midrule
Total & 373,920 & 2,174,211 & 6,948,950 & 4,606,940\\
\bottomrule 

%%%%%%%%%%%%%%%%%%%%%% LLaVAR
\hline
\hline
\toprule
\multicolumn{5}{c}{\textbf{LLaVAR~\cite{llavar}}} \\
\midrule
% \textbf{Dataset Group} & \textbf{\# Images} & \textbf{\# Captions} & \textbf{\# Bounding Boxes} & \textbf{\# QA} \\
% \midrule

LAION-5B\cite{laion-5b} & 13,104 & 13,104 & 0 & 0\\
EasyOCR\cite{easyocr} (our work) & 13,104 & 0 & 49,282 & 0\\
\midrule
Total & 13,104 & 13,104 & 49,282 & 0\\
\bottomrule

%%%%%%%%%%%%%%%%%%%%%% 1x Mix
\hline
\hline
\toprule
\multicolumn{5}{c}{\textbf{1x Mix}} \\
\midrule
% \textbf{Dataset Group} & \textbf{\# Images} & \textbf{\# Captions} & \textbf{\# Bounding Boxes} & \textbf{\# QA} \\
% \midrule

A-OKVQA\cite{a-okvqa} & 8,195 & 0 & 0 & 8,195\\
COCO Captions\cite{cococaption} & 56,979 & 285,058 & 424,259 & 0\\
InfographicVQA\cite{infographicvqa} & 319 & 0 & 0 & 1,735\\
Localized Narratives\cite{localizednarratives} & 56,979 & 64,983 & 0 & 0\\
LVIS\cite{lvis} & 47,546 & 94,770 & 603,979 & 0\\
OCR-VQA\cite{ocrvqa} & 60,418 & 177,434 & 0 & 292,497\\
OK-VQA\cite{okvqa} & 4,514 & 0 & 0 & 4,514\\
refCOCO\cite{refcoco} & 8,692 & 0 & 61,943 & 0\\
refCOCOg\cite{refcoco} & 11,092 & 0 & 40,827 & 0\\
refCOCO+\cite{refcoco} & 8,691 & 0 & 61,810 & 0\\
TextVQA\cite{textvqa} & 11,497 & 11,497 & 0 & 18,138\\
TextCaps\cite{textcaps} & 11,497 & 631,955 & 0 & 0\\
Visual7w\cite{visual7w} & 7,484 & 0 & 0 & 36,333\\
Visual Genome\cite{visual_genome} & 55,951 & 0 & 3,169,962 & 0\\
Visual Spatial Reasoning\cite{vsr} & 2,972 & 5,293 & 0 & 0\\

\midrule
Total & 160,643 & 1,270,990 & 4,362,780 & 361,412\\
\bottomrule

%%%%%%%%%%%%%%%%%%%%%% DocOwl
\hline
\hline
\toprule
\multicolumn{5}{c}{\textbf{DocOwl~\cite{mplug_docowl}}} \\
\midrule
% \textbf{Dataset Group} & \textbf{\# Images} & \textbf{\# Captions} & \textbf{\# Bounding Boxes} & \textbf{\# QA} \\
% \midrule

ChartQA\cite{masry-etal-2022-chartqa} & 17,023 & 0 & 0 & 26,356\\
DocVQA\cite{mathew2021docvqa} & 8,675 & 0 & 0 & 43,963\\
DocOwl\cite{mplug_docowl} & 50,999 & 0 & 761,060 & 0\\
InfographicVQA\cite{infographicvqa} & 1,945 & 0 & 0 & 9,190\\
VisualMCR\cite{machinereadingcomprehension} & 2,796 & 0 & 0 & 11,016\\
TextVQA\cite{textvqa} & 20,561 & 20,561 & 0 & 32,426\\

\midrule
Total & 51,000 & 20,561 & 761,060 & 122,951\\
\bottomrule

%%%%%%%%%%%%%%%%%%%%%% Shikra
% \hline
% \hline
% \toprule
% \multicolumn{5}{c}{\textbf{Shikra~\cite{shikra}}} \\
% \midrule
% % \textbf{Dataset Group} & \textbf{\# Images} & \textbf{\# Captions} & \textbf{\# Bounding Boxes} & \textbf{\# QA} \\
% % \midrule

% Flickr 30K\cite{flickr30k} & 14,923 & 74,614 & 0 & 0\\
% refCOCO\cite{refcoco} & 4,638 & 0 & 31,218 & 0\\
% refCOCOg\cite{refcoco} & 6,487 & 0 & 22,756 & 0\\
% refCOCO+\cite{refcoco} & 4,638 & 0 & 31,245 & 0\\
% Visual7w\cite{visual7w} & 1,929 & 0 & 0 & 9,791\\
% Visual Genome\cite{visual_genome} & 15,480 & 0 & 469,580 & 0\\
% VQA V2\cite{vqav2} & 25,947 & 0 & 0 & 142,076\\

% \midrule
% Total & 52,123 & 74,614 & 554,799 & 151,867\\
% \bottomrule

\end{tabular}

% \vspace{0.2cm}

% \begin{tabular}{cccccccccc}
% % \hline
% \toprule
% \multicolumn{5}{c}{\textbf{LLaVAR}} \\
% \midrule
% \textbf{Dataset Group} & \textbf{\# Images} & \textbf{\# Captions} & \textbf{\# Bounding Boxes} & \textbf{\# QA} \\
% \midrule

% llavar & 13,104 & 13,104 & 49,282 & 0\\

% \midrule
% Total & 13,104 & 13,104 & 49,282 & 0\\
% \bottomrule

% \end{tabular}

\label{table:DatasetGroup}
\end{table*}

% \input{figure_tex/supp_prompt_shikra}
% \subsubsection{Data}
% % Placeholder for data details
% % Add additional information here if needed

\clearpage

{
    \small
    \bibliographystyle{ieeenat_fullname}
    \bibliography{main}
}

\end{document}